\documentclass[a4paper,fleqn]{cas-dc}

\usepackage[numbers]{natbib}
\usepackage{amsmath,amssymb}
\usepackage{pifont}
\usepackage{diagbox}
\usepackage{arydshln}
\usepackage[linesnumbered,ruled,vlined]{algorithm2e}
\usepackage{subcaption}
\usepackage{multirow}
\usepackage{graphicx}
\usepackage{adjustbox}
\usepackage{makecell}
\usepackage{xcolor}
\usepackage{booktabs}

\begin{document}
\let\WriteBookmarks\relax
\def\floatpagepagefraction{1}
\def\textpagefraction{.001}

\shorttitle{CASO PAD}
\shortauthors{S. Khan}

\title[mode=title]{Face Presentation Attack Detection via Content-Adaptive Spatial Operators}

\author[1,2]{Shujaat Khan}
\ead{shujaat.khan@kfupm.edu.sa}
\cormark[1]
\cortext[cor1]{Corresponding author}

\affiliation[1]{organization={Department of Computer Engineering, College of Computing and Mathematics, King Fahd University of Petroleum \& Minerals},
    city={Dhahran},
    postcode={31261},
    country={Saudi Arabia}}
\affiliation[2]{organization={SDAIA--KFUPM Joint Research Center for Artificial Intelligence, King Fahd University of Petroleum \& Minerals},
    city={Dhahran},
    postcode={31261},
    country={Saudi Arabia}}


\begin{abstract}
Face presentation attack detection (FacePAD) is critical for securing facial authentication against print, replay, and mask-based spoofing. This paper proposes CASO-PAD, an RGB-only, single-frame model that enhances MobileNetV3 with content-adaptive spatial operators (involution) to better capture localized spoof cues. Unlike spatially shared convolution kernels, the proposed operator generates location-specific, channel-shared kernels conditioned on the input, improving spatial selectivity with minimal overhead. CASO-PAD remains lightweight (3.6M parameters; 0.64 GFLOPs at $256\times256$) and is trained end-to-end using a standard binary cross-entropy objective. Extensive experiments on Replay-Attack, Replay-Mobile, ROSE-Youtu, and OULU-NPU demonstrate strong performance, achieving 100/100/98.9/99.7\% test accuracy, AUC of 1.00/1.00/0.9995/0.9999, and HTER of 0.00/0.00/0.82/0.44\%, respectively. On the large-scale SiW-Mv2 Protocol-1 benchmark, CASO-PAD further attains 95.45\% accuracy with 3.11\% HTER and 3.13\% EER, indicating improved robustness under diverse real-world attacks. Ablation studies show that placing the adaptive operator near the network head and using moderate group sharing yields the best accuracy--efficiency balance. Overall, CASO-PAD provides a practical pathway for robust, on-device FacePAD with mobile-class compute and without auxiliary sensors or temporal stacks.
\end{abstract}


\maketitle
\section{Introduction}
Facial recognition is now integral to mobile payments, surveillance, border control, and personal device authentication. Its appeal lies in being non-intrusive and accurate, powering applications such as smartphone unlocking \cite{wang2023dynamic}, mobile payments \cite{ye2023mobile}, surveillance systems \cite{fang2023surveillance}, access control \cite{lee2020face}, and attendance monitoring \cite{alhanaee2021face}. Leveraging large datasets and high-performance computing, modern recognition systems have surpassed 99\% accuracy \cite{grother2019face}. Yet, this very sensitivity to facial data introduces vulnerabilities to spoofing attacks, where adversaries use counterfeit inputs such as printed photos, display replays, or 3D masks \cite{galbally2014biometric}. As adoption expands into sensitive domains like finance and healthcare, the need for reliable anti-spoofing solutions has become paramount.

FacePAD systems aim to distinguish genuine (\emph{bonafide}) inputs from spoof attempts, ensuring the security of facial recognition technologies \cite{wen2015face}. Applications extend beyond traditional security to e-commerce, mobile device authentication, and remote access and the demand for robust and deployment-friendly PAD methods has never been greater \cite{yu2022deep}.

Early approaches to FacePAD relied on handcrafted features such as SIFT \cite{lowe2004distinctive}, SURF \cite{bay2006surf}, and HOG \cite{dalal2005histograms}, often combined with SVMs or LDA classifiers. Texture- and motion-based cues, including eye blinks and lip motion, also proved useful \cite{pan2007eyeblink, komulainen2013context}. However, these techniques struggled under variable lighting, backgrounds, and camera qualities, limiting their real-world applicability.

Beyond basic photo or video replays, adversaries now employ sophisticated techniques, including high-fidelity 3D masks, adversarial perturbations, and machine learning-driven attacks \cite{sun2020face, szegedy2013intriguing} and the arms race between attackers and FacePAD continues. Although data augmentation and unified frameworks such as UniFAD \cite{deb2023unified} have been proposed, they often require large-scale datasets and remain limited to specific attack scenarios, especially under challenging white-box conditions. Compounding the difficulty, no benchmark dataset can encompass the diversity of possible attack types \cite{zhang2012face}.

To address data scarcity, generative models such as GANs have been explored for synthesizing spoofing samples \cite{denton2015deep, radford2015unsupervised}. Recent works demonstrate that adversarial training with synthetic data can improve PAD robustness \cite{siddique2025advspoofguard}, though challenges of model scalability and efficiency remain. Deep learning, particularly Convolutional Neural Networks (CNNs), has since transformed FacePAD by capturing subtle, discriminative features. CNN-based models perform well but often their high computational cost hinders real-time use in resource-limited environments such as mobile devices. 

To deal with the challenges of FacePAD, we propose a content-adaptive framework for resource-constrained environments. The proposed network utilizes a computationally efficient backbone architecture specifically designed for edge devices \cite{howard2019searching}. The major contributions of the proposed system lie in its \emph{content-adaptive spatial operators} mechanism, which fuses feature extraction with novel group-wise involution \cite{li2021involution} signal processing to detect input-specific, fine-grained spoofing cues effectively. Unlike fixed depthwise filters, the proposed approach generates location-specific, channel-shared kernels conditioned on the input, thereby enhancing spatial adaptivity and improving discrimination of spoof artifacts. The contributions of this work are as follows:

\begin{itemize}
    \item A RGB single-frame FacePAD model, CASO-PAD, that augments MobileNetV3 \cite{howard2019searching} with content-adaptive involution \cite{li2021involution} layers at selected stages for enhanced spoof detection.
    \item A detailed audit of learned kernels and their interpretation for model explainability.
    \item A streamlined training setup with ablations over model architectures, kernel size, placement strategy, image size, and group sharing for efficient learning.
    \item Comprehensive evaluation on Replay-Attack \cite{chingovska2012effectiveness}, Replay-Mobile \cite{costa2016replay}, OULU-NPU \cite{boulkenafet2017oulu}, ROSE-Youtu \cite{Li2018UnsupervisedDA}, and SiW-Mv2 \cite{xiaoguo2022MDFAS} datasets using standard metrics (e.g., Accuracy, AUC, EER, HTER, APCER, BPCER, and ACER, etc), showing strong HTER--efficiency trade-offs against baselines.
\end{itemize}

The remainder of this paper is structured as follows: Section \ref{sec:literature_review} reviews existing FacePAD methods and their limitations. Section \ref{sec:proposed_method} details the proposed framework. Section \ref{sec:results_discussion} discusses findings and comparisons, while Section \ref{sec:kernel_audit} and \ref{sec:ablation} present the kernel audit and detailed ablation studies respectively. Finally Section \ref{sec:conclusion} concludes the study.

\section{Related Work} \label{sec:literature_review}

FacePAD has attracted substantial interest, producing a broad taxonomy of methods: (i) \emph{spatial/appearance} techniques that analyze frame-level texture and color cues; (ii) \emph{temporal} approaches that exploit motion and dynamics; and (iii) \emph{multimodal} methods that incorporate auxiliary signals such as Remote photoplethysmography (rPPG), depth, thermal, or infrared. Recent work also explores adversarial robustness and domain generalization. Below, we synthesize these directions and highlight gaps motivating an efficient, RGB-only solution.

\subsection{Spatial/Appearance Methods}

\paragraph{Classical handcrafted features.}
Early PAD relied on handcrafted descriptors to capture fine-grained textural differences between bonafide and spoof images. Common choices include Local Binary Patterns (LBP) \cite{ojala2002multiresolution}, Histograms of Oriented Gradients (HOG) \cite{dalal2005histograms}, and SIFT \cite{lowe2004distinctive}. M\"{a}\"{a}tt\"{a} et al.\ \cite{maatta2011face} used LBP to expose print-induced artifacts, while other works explored color spaces (e.g., HSV, YCbCr) and frequency cues to better separate spoof patterns from genuine skin reflectance \cite{boulkenafet2016face, boulkenafet2015face, li2004live}. Although these methods are effective against simple photo attacks, they degrade under varied illuminations, cameras, and high-quality 3D mask or replay attacks \cite{patel2016secure}.

\paragraph{Deep learning and hybrid models.}
CNN-based FacePAD superseded handcrafted approaches by learning discriminative features end-to-end. Representative advances include two-stream or attention mechanisms over RGB and illumination-invariant spaces, pixel-level supervision for local spoof cues \cite{sun2020face}, and feature fusion with learned or autoencoder-derived representations \cite{rehman2019face}. Additional strands investigate adversarial robustness \cite{bousnina2021unraveling}, one-class client-specific modeling \cite{fatemifar2021client}, domain adaptation/generalization \cite{wang2020unsupervised, wang2022domain}, and diffusion-aided hybrids \cite{alassafi2023novel}. Hybrid methods deliberately combine handcrafted texture with CNNs or integrate classical features to improve cross-dataset generalization \cite{patel2016cross}. While these designs often boost accuracy, many are computationally expensive and less suitable for real-time deployment on edge devices.

\subsection{Temporal and Motion-Based Methods}

Temporal information helps reveal inconsistencies characteristic of replay and mask attacks. Classic works used optical flow or motion patterns \cite{bao2009liveness} and eye-blink analysis via undirected conditional random fields (CRFs)  \cite{pan2007eyeblink}. With deep learning, CNN-RNN hybrids (e.g., CNN+LSTM) capture spatiotemporal dependencies, while deep dynamic textures and flow-guided models further enhance temporal sensitivity \cite{shao2018joint}. More recent efforts compress or summarize clip dynamics through temporal sequence sampling and stacked recurrent encoders \cite{muhammad2022self, dhiman2024deep}. Despite strong accuracy, these approaches are slow in nature due to high latency and can also be parameter- and compute-intensive, limiting use on mobile hardware.

\subsection{Multimodal Signals: rPPG, Depth, and Infrared}

The rPPG leverages subtle color changes induced by blood flow as a liveness prior. Methods range from correspondence features and noise-aware templates to transformer-based encoders \cite{Liu_2018_ECCV, 9316990, yu2021transrppg}. rPPG can separate bonafide from spoofed content and has been applied to 3D mask scenarios \cite{liu20163d}, but it is sensitive to illumination, motion, and video quality, often requiring longer capture windows \cite{yao2021rppg, liu2020temporal}. Beyond rPPG, multispectral and infrared imaging capture non-visible cues and temperature or material differences \cite{kotwal2019multispectral, li2024face}. These systems are powerful but typically demand specialized hardware, limiting scalability in low-cost deployments.

\subsection{Summary and Motivation}

Classical texture and motion pipelines are fast but fragile under domain shifts and sophisticated spoofs \cite{patel2016secure}. Deep models substantially improve accuracy, yet many require heavy backbones, auxiliary sensors, or complex training that impedes real-time use \cite{alassafi2023novel}. Multimodal signals (rPPG, depth, thermal/IR) enhance robustness but demand controlled capture conditions or specialized hardware \cite{liu20163d, kotwal2019multispectral, li2024face}. Temporal encoders strengthen replay/mask detection \cite{shao2018joint, muhammad2022self, dhiman2024deep}, but LSTM/transformer stacks are often parameter-intensive for phones and embedded devices.

These trade-offs motivate \emph{efficient, RGB-only} models that retain strong discriminative power while meeting edge constraints. Lightweight backbones such as MobileNetV2/V3 and ShuffleNet achieve favorable accuracy--efficiency via depthwise separable convolutions and attention; however, their spatially shared kernels can under-capture localized spoof artifacts. To address this, \emph{involution} \cite{li2021involution} replaces channel-specific, spatially shared kernels with \emph{spatially specific, channel-shared} kernels produced by a lightweight generator, enabling content-adaptive filtering with low overhead. 


\section{Proposed Method}\label{sec:proposed_method}

This section outlines the datasets employed in this study, the performance evaluation criteria, the architecture of the proposed network, and the adopted training protocol.

\subsection{Datasets}

To assess the robustness and generalization capability of the proposed approach, five well-established face anti-spoofing datasets are used: Replay-Attack \cite{chingovska2012effectiveness}, Replay-Mobile \cite{costa2016replay}, OULU-NPU \cite{boulkenafet2017oulu}, ROSE-Youtu \cite{Li2018UnsupervisedDA}, and SiW-Mv2 \cite{xiaoguo2022MDFAS}. A concise summary of their characteristics is provided in Table~\ref{tab:datasetsummary}.

\begin{table}[t]
    \centering
    \caption{Overview of the benchmark datasets used for evaluation.}
    \resizebox{\linewidth}{!}{
    \begin{tabular}{l|c|c|c}
        \hline
        \textbf{Dataset} & \textbf{Release Year} & \textbf{Participants} & \textbf{Authentic / Attack Samples} \\
        \hline
        Replay-Attack \cite{chingovska2012effectiveness} & 2012 & 50 & 300 / 1000 \\
        Replay-Mobile \cite{costa2016replay} & 2016 & 40 & 550 / 640 \\
        OULU-NPU \cite{boulkenafet2017oulu} & 2017 & 55 & 990 / 3,960 \\
        ROSE-Youtu \cite{Li2018UnsupervisedDA} & 2018 & 20 & 1000 / 2350 \\
        SiW-Mv2 \cite{xiaoguo2022MDFAS} & 2022 & 600 & 785 / 915 \\
        \hline
    \end{tabular}}
    \label{tab:datasetsummary}
\end{table}

\begin{figure}[!h]
    \centering
    \includegraphics[width=0.95\linewidth]{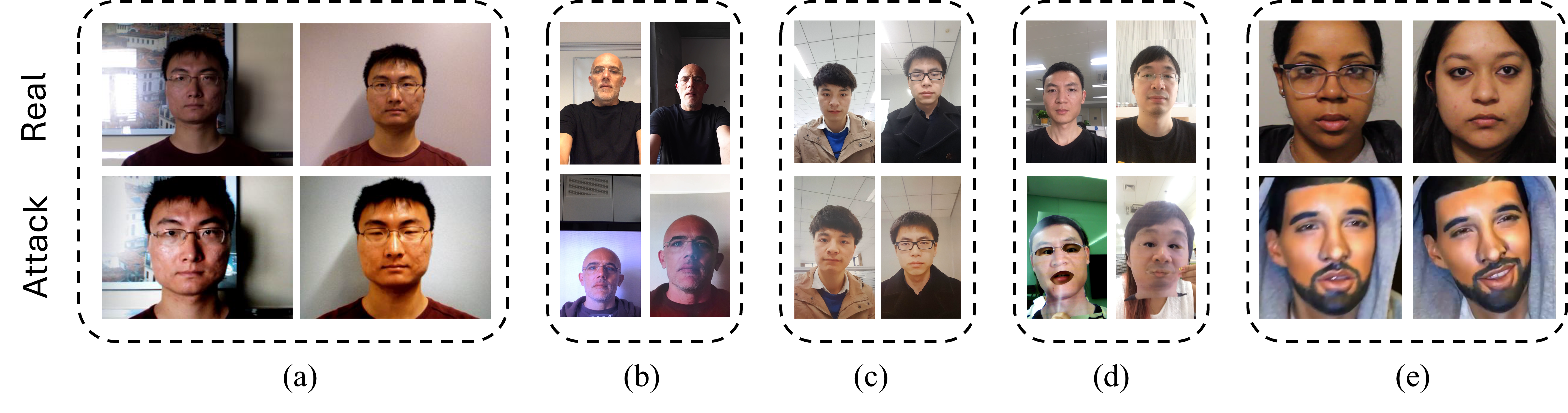}
    \caption{Sample frames from authentic and spoofed videos in (a) RA, (b) RM, (c) OULU-NPU, (d) RY, and (e) SiW-Mv2 datasets. Genuine samples appear in the top row, while the bottom row shows different attack types.}
    \label{fig:dataset}
\end{figure}

\subsubsection{Replay-Attack}

The Replay-Attack dataset \cite{chingovska2012effectiveness} comprises 1,300 video sequences encompassing both genuine and attack scenarios (including photos and video replays). The data were collected from 50 individuals under both controlled and variable illumination using different cameras. The dataset is partitioned into disjoint training, development, and testing subsets, ensuring that no subject appears in more than one split. Figure~\ref{fig:dataset}(a) presents illustrative frames from genuine and spoofed sequences.

\subsubsection{Replay-Mobile}

The Replay-Mobile dataset \cite{costa2016replay} was curated to evaluate mobile-based face recognition and presentation attack detection systems. It consists of 1,190 video samples from 40 participants, captured in diverse lighting settings using smartphone cameras. The attacks include both printed photo and digital replay types. Example genuine and spoof frames are shown in Figure~\ref{fig:dataset}(b).

\subsubsection{OULU-NPU}
The OULU-NPU \cite{boulkenafet2017oulu} dataset consists of $4{,}950$ video samples collected from $55$ subjects using six mobile devices (HTC Desire Eye, ASUS Zenfone Selfie, Oppo N3, Meizu MX5, Sony Xperia C5 Ultra Dual, and Samsung Galaxy S6 Edge). The data were acquired across three sessions under varying background scenes and illumination conditions. Presentation attacks were generated using two print and two display devices. The dataset is partitioned subject-disjointly into training and testing sets with $20$ subjects each, while the remaining $15$ subjects are reserved for validation. Example genuine and spoof frames are shown in Figure~\ref{fig:dataset}(c),

\subsubsection{ROSE-Youtu}

The ROSE-Youtu dataset \cite{Li2018UnsupervisedDA} includes 3,350 videos from 20 subjects, recorded under multiple illumination conditions and using different camera models. It contains three distinct spoof types: video replay, printed photo, and paper mask attacks. As illustrated in Figure~\ref{fig:dataset}(d), this dataset poses a considerable challenge for FacePAD research due to its diverse acquisition conditions and attack modalities.

\subsubsection{Spoofing in the Wild (SiW-Mv2) Dataset}
SiW-Mv2 \cite{xiaoguo2022MDFAS} is a large-scale FacePAD benchmark designed to evaluate robustness under a wide variety of spoofing scenarios, comprising 14 different presentation attack categories. The dataset includes 915 spoof videos collected from 600 subjects, along with 785 bona fide recordings from 493 individuals. The attack set covers diverse and challenging conditions such as replay-based attacks, partial manipulation of facial regions (e.g., eye-only spoofs), silicone mask presentations, and paper-based artifacts. Representative sample frames from SiW-Mv2 are illustrated in Figure~\ref{fig:dataset}(e).

\subsection{Pre-processing}
In this work, the pre-processing stage involves a center-cropping strategy designed to retain the natural aspect ratio of each frame while avoiding geometric distortion during resizing. For both the training and testing phases, the same procedure is applied to ensure consistency.

The adaptive cropping operation follows the steps below:
\begin{itemize}
    \item Determine the smaller dimension $\rho$ between the frame height $(M)$ and width $(N)$.
    \item Extract a centered square region of size $(\rho \times \rho)$ from the original frame.
\end{itemize}

\begin{figure}[ht]
    \centering
    \includegraphics[width=\linewidth]{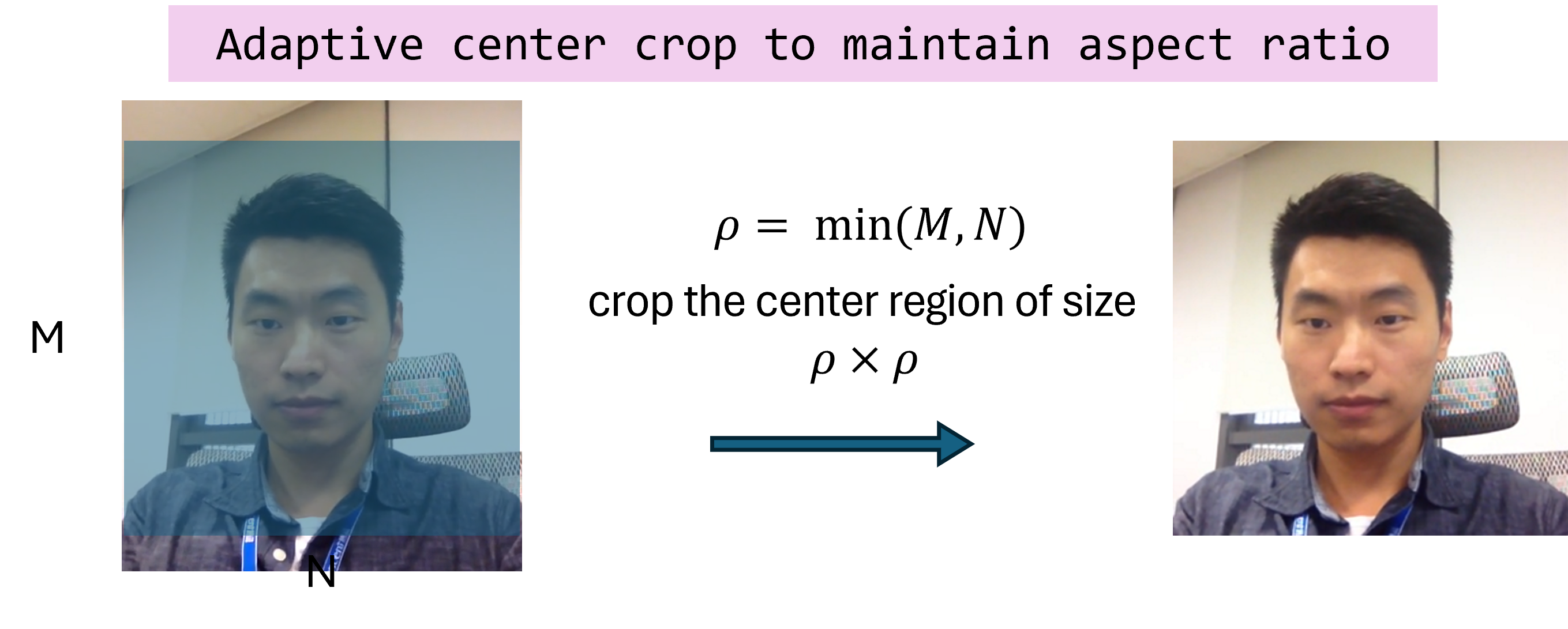}
    \caption{Illustration of the adaptive center-cropping process used in the pre-processing pipeline. This operation ensures consistent framing and preserves the original aspect ratio across all samples.}
    \label{fig:center_crop}
\end{figure}

Figure~\ref{fig:center_crop} illustrates the adaptive center-cropping procedure implemented in this study, which serves as the foundation for subsequent data augmentation and normalization.

\subsection{Proposed Network Architecture}

In CNNs, a convolution layer typically apply a single spatial kernel uniformly across the feature map. This entails two well-known constraints:
\begin{enumerate}
    \item \textbf{Spatially agnostic kernels:} the same filter is reused at every location, limiting sensitivity to local texture, scale, and structure.
    \item \textbf{Fixed cross-channel coupling:} channel interactions are prescribed by the learned $C'\!\times\!C$ kernel and remain independent of spatial context.
\end{enumerate}

Before contrasting alternative operators, we fix notation and the receptive-field convention used throughout.
\paragraph{Notation.}
Let $X \in \mathbb{R}^{H \times W \times C}$ denote the input, $Y \in \mathbb{R}^{H \times W \times C'}$ the output, and let $\Omega=\{(u,v)\,:\,|u|\le \lfloor k/2\rfloor,\,|v|\le \lfloor k/2\rfloor\}$ be the $k{\times}k$ receptive field.

With this notation, the conventional convolution layer can be written as:
\subsubsection{Standard Convolution}
\begin{equation}
Y(i,j,c_o) \;=\; \sum_{c_i=1}^{C} \;\sum_{(u,v)\in\Omega} K(c_o,c_i,u,v)\; X(i{+}u, j{+}v, c_i),
\end{equation}
where a single kernel $K$ is shared across all spatial positions $(i,j)$.

While standard convolution mixes all channels everywhere, many efficient backbones restrict cross-channel mixing to subsets to reduce cost. This yields group (and in the extreme, depthwise) convolution:
\subsubsection{Group Convolution}
Partition channels into $G$ disjoint groups of size $S=C/G$. Mixing occurs only within the same group:
\begin{equation}
Y(i,j,c_o) \;=\; \sum_{c_i \in \mathcal{G}(c_o)} \;\sum_{(u,v)\in\Omega} K_g(c_o,c_i,u,v)\; X(i{+}u, j{+}v, c_i),
\end{equation}
where $g=g(c_o)\!\in\!\{1,\dots,G\}$ indexes the group of $c_o$ and $\mathcal{G}(c_o)$ is its input-channel set. Standard convolution is $G{=}1$; depthwise convolution corresponds to $G{=}C$.

Both of the above still reuse the \emph{same} spatial kernel at every location, which dilutes sensitivity to position-dependent artifacts typical in FacePAD (e.g., specular highlights or print borders). To address this, we adopt a location-adaptive operator.

\subsubsection{Involution (location-adaptive, channel-shared)}
Involution~\cite{li2021involution} replaces a \emph{spatially invariant} kernel with \emph{location-specific} kernels $H(i,j,u,v)$ that are shared across channels:
\begin{equation}
Y(i,j,c) \;=\; \sum_{(u,v)\in\Omega} H(i,j,u,v)\; X(i{+}u, j{+}v, c),
\end{equation}
where $H$ is generated on-the-fly from features (e.g., via a compact kernel-generator network).

Involution preserves the low compute of depthwise/grouped designs while allowing the spatial kernel to vary across $(i,j)$, a property we will exploit in our content-adaptive head.

\subsubsection{Proposed Group involution (GI)}
Building on the advantages of involution, we introduce an adaptive group involution (GI). The proposed operator is \emph{group-wise} and \emph{location-adaptive}. For $G$ groups, where one spatial kernel per group and location is generated; that kernel is \emph{shared across channels} within the group and applied depthwise (no channel mixing in the spatial op):
\begin{equation}
Y(i,j,c) \;=\; \sum_{(u,v)\in\Omega} H\big(i,j,g(c),u,v\big)\; X(i{+}u, j{+}v, c),
\label{eq:agi}
\end{equation}
where $g(c)\in\{1,\dots,G\}$ maps channel $c$ to its group. Cross-channel interaction is provided by surrounding $1{\times}1$ pointwise layers (squeeze/expand), while the spatial operator itself remains group-shared and content-adaptive.

\paragraph{Remarks.}
\begin{itemize}
    \item \textbf{Special cases:} $G{=}C$ reduces to depthwise involution \cite{li2021involution} (one kernel per channel); $G{=}1$ reduces to channel-shared involution\cite{li2021involution}.
    \item \textbf{Efficiency:} the spatial application in~\eqref{eq:agi} scales as $\mathcal{O}(C\,k^2HW)$, while the kernel-generator cost scales with group count $G$ and the reduction ratio in the $1{\times}1$ bottleneck; both are lightweight compared to full $C{\times}C'$ convolution.
\end{itemize}

The proposed GI provides a flexible trade-off between expressivity and efficiency by (i) adapting kernels to \emph{spatial content} and (ii) controlling \emph{channel sharing} via the number of groups $G$. This is especially beneficial for lightweight backbones (e.g., MobileNetV3) and tasks demanding fine spatial awareness (e.g., face PAD, texture analysis, medical imaging), where directional, content-adaptive filtering improves selectivity without incurring the cost of full $C{\times}C'$ spatial mixing.

\subsubsection{Proposed MobV3-GI: Content-Adaptive Spatial Operator-based Network}
\begin{figure*}[ht]
    \centering
    \includegraphics[width=\textwidth]{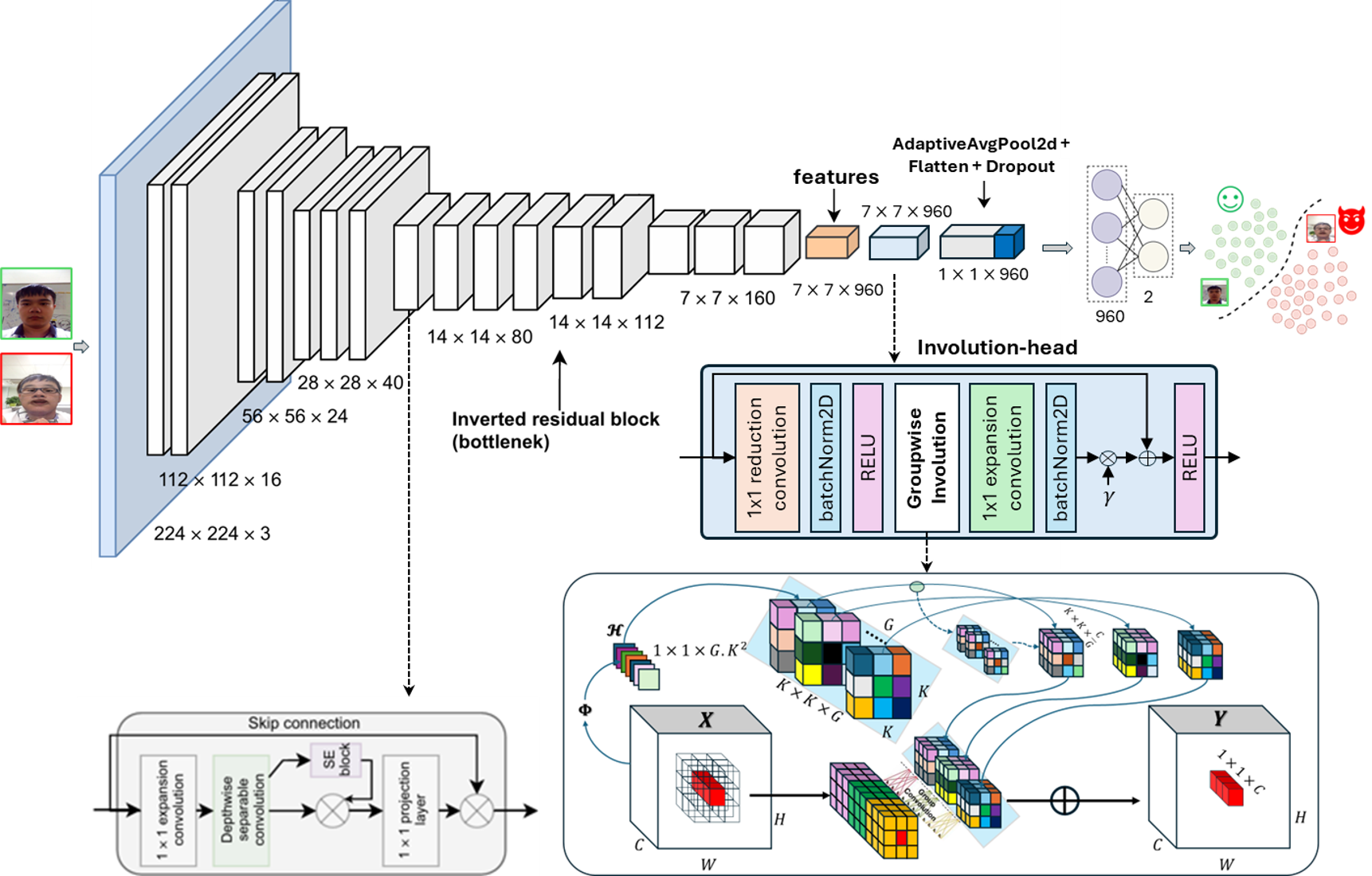}
    \caption{Schematic diagram of the proposed content adaptive spatial operator-based deep learning network.}
    \label{fig:arch_caso_pad}
\end{figure*}

The proposed network architecture (as shown in Figure~\ref{fig:arch_caso_pad}) begin with MobileNetV3-Large~\cite{howard2019searching}. A typical block comprises expansion ($1{\times}1$), depthwise $3{\times}3$, squeeze-and-excitation (SE), and projection ($1{\times}1$). Herein, selected \emph{depthwise} convolutions are replaced by proposed \emph{group involution}:
\begin{equation}
\mathbf{Y}(i,j,:) \;=\; \sum_{(u,v)\in\Omega_k}\! \mathcal{K}_{i,j}(u,v)\,\mathbf{X}(i{+}u,\,j{+}v,:),
\end{equation}
where $\mathcal{K}_{i,j}\!\in\!\mathbb{R}^{k\times k}$ is spatially varying and channel-shared. Kernels are generated by a lightweight function $g_\phi$ (e.g., $1{\times}1$ convs + BN + nonlinearity) from a squeezed version of $\mathbf{X}$. Group sharing ($G$) reduces cost by generating kernels per channel-group.

Global average pooling yields $\mathbf{z}\in\mathbb{R}^{d}$; a linear head produces logits $\mathbf{s}=W\mathbf{z}+b\in\mathbb{R}^{2}$. We optimize cross-entropy with optional label smoothing ($\epsilon{=}0.05$):

\paragraph{Placement.}
To balance accuracy and efficiency, involution/GI is applied at both low- and high-resolution stages; Section~\ref{sec:ablation} ablates early vs.\ late placement.

\begin{algorithm} 
\caption{} \label{alg:proposed}
\SetAlgoLined
\KwIn{Batch of videos \( X \in \mathbb{R}^{B \times C \times H \times W} \)}
\KwOut{Class predictions \( y \in \mathbb{R}^{B \times \text{num\_classes}} \)}

\emph{Initialize Model:}\\
\quad Load MobV3-GI as the backbone CNN\\
\emph{Frame-Level Feature Extraction:}\\
\quad Extract features \( \mathbf{f} = \text{CNN}(X[:, :, :, :]) \in \mathbb{R}^{B \times 960} \)\\

\emph{Fully Connected Layer:}\\
\quad Classify: \( y = \text{FC}(\mathbf{F}_{\text{avg}}) \)

\Return{Class predictions \( y \)}
\end{algorithm}

Algorithm~\ref{alg:proposed} summarizes the overall inference workflow of the proposed CASO-PAD system. 
It outlines how the MobV3-GI backbone extracts spatially adaptive features from each input frame, and performs classification via a lightweight fully connected head. 
The algorithmic outline emphasizes the model’s simplicity and computational efficiency, making it easily reproducible for mobile and embedded implementations.

\subsection{Training Setup}

The training of the proposed model is performed using the Adam optimizer with a learning rate of \(10^{-4}\). The optimization objective is the binary cross-entropy (BCE) loss, formulated as:

\begin{equation}
    L = -\frac{1}{N} \sum_{i=1}^{N} \big[ y_i \log(p_i) + (1 - y_i)\log(1 - p_i) \big],
\end{equation}

where \(y_i\) denotes the ground-truth label and \(p_i\) represents the model’s estimated probability for the genuine (live) class.

Training is conducted for $100$ epochs using mini-batches of size $32$. Early stopping with a patience threshold of $5$ epochs is employed to prevent overfitting, based on the validation loss trend. During the testing phase, inference is performed with a batch size of $256$ to ensure computational efficiency.

All experiments are implemented in the PyTorch framework and executed on an NVIDIA GeForce RTX 5080 GPU equipped with 16~GB of VRAM. Unless otherwise stated, all input frames are resized to a spatial resolution of $256 \times 256$ pixels. The source code and implementation details are publicly available at:  
\url{https://github.com/Shujaat123/CASO-PAD}.

\section{Results and Discussion}\label{sec:results_discussion}

This section presents a detailed evaluation of the proposed method across standard benchmark datasets and multiple performance metrics. We first discuss the quantitative outcomes obtained on Replay-Attack (RA), Replay-Mobile (RM), OULU-NPU, ROSE-Youtu (RY), and SiW-Mv2, followed by a comparative analysis with existing state-of-the-art FacePAD approaches.

\subsection{Performance Evaluation}
The proposed \emph{Content-Adaptive Spatial Operators for FacePAD} (CASO-PAD) is built on an involution-augmented MobileNetV3 backbone. Table~\ref{tab:performance} reports the overall performance under the default configuration ($256\times256$ input resolution, $G{=}120$ groups, reduction factor $r{=}4$, and kernel size $k{=}5$). Across five benchmarks, CASO-PAD demonstrates strong discriminative capability, achieving perfect or near-perfect separation on controlled datasets and maintaining robust performance on the more challenging in-the-wild protocol.

\begin{table}[!ht]
    \centering
    \caption{Performance of the proposed method on Replay-Attack (RA), Replay-Mobile (RM), OULU-NPU, ROSE-Youtu (RY), and SiW-Mv2 (Protocol-1) datasets. Results are reported as mean$\pm$std over 3 runs where applicable.}
    \resizebox{\linewidth}{!}{
    \begin{tabular}{l | c c c c c}
        \hline
        \textbf{{\diagbox[width=14em]{Metric}{Dataset}}} 
        & \textbf{RA} & \textbf{RM} & \textbf{OULU} & \textbf{RY} & \textbf{SiW-Mv2} \\
        \hline
        Test Accuracy (\%) 
        & 100$\pm$0.0 & 100$\pm$0.0 & 99.68$\pm$0.13 & 98.90$\pm$0.35 & 95.45$\pm$1.63 \\
        Youden Index (YI) 
        & 1.0$\pm$0.0 & 1.0$\pm$0.0 & 0.991$\pm$0.001 & 0.98$\pm$0.0 & 0.938$\pm$0.020 \\
        AUC-ROC 
        & 1.0$\pm$0.0 & 1.0$\pm$0.0 & 0.9999$\pm$0.0000 & 0.99$\pm$0.0 & 0.9906$\pm$0.0022 \\
        EER (\%) 
        & 0.0$\pm$0.0 & 0.0$\pm$0.0 & 0.44$\pm$0.11 & 0.82$\pm$0.09 & 3.13$\pm$0.70 \\ 
        FAR (\%) 
        & 0.0$\pm$0.0 & 0.0$\pm$0.0 & 0.23$\pm$0.24 & 0.82$\pm$0.09 & 2.76$\pm$0.55 \\ 
        HTER (\%) 
        & 0.0$\pm$0.0 & 0.0$\pm$0.0 & 0.44$\pm$0.04 & 0.82$\pm$0.21 & 3.11$\pm$1.02 \\ 
        FRR (\%) 
        & 0.0$\pm$0.0 & 0.0$\pm$0.0 & 0.65$\pm$0.26 & 0.82$\pm$0.34 & 3.45$\pm$1.75 \\ 
        \hline
    \end{tabular}}
    \label{tab:performance}
\end{table}

On Replay-Attack (RA) and Replay-Mobile (RM), CASO-PAD achieves flawless separation between bona fide and spoof classes, reaching 100\% accuracy and AUC-ROC of 1.0 with 0.0\% EER/HTER. These results indicate strong generalization under the controlled capture conditions and device variations typical of RA/RM.  

On OULU-NPU, which introduces larger variability in illumination, background, capture devices, and attack instruments, CASO-PAD maintains near-ceiling performance (99.68\% accuracy, AUC-ROC 0.9999) with low error rates (EER 0.44\%, HTER 0.44\%).  

ROSE-Youtu (RY) is more challenging due to diverse spoof types (print, replay, and mask/paper-mask) and heterogeneous acquisition settings. CASO-PAD sustains strong robustness with 98.90\% accuracy, AUC-ROC 0.99, and EER/HTER of 0.82\%, while preserving balanced sensitivity and specificity (Youden Index 0.98).  

Finally, on SiW-Mv2 Protocol-1, which reflects large-scale, in-the-wild spoofing conditions with substantial subject and attack diversity, CASO-PAD achieves 95.45\% accuracy and AUC-ROC 0.9906 with 3.11\% HTER and 3.13\% EER. Overall, these results confirm that the proposed content-adaptive spatial operator improves discriminability on standard benchmarks while remaining robust under more realistic, unconstrained attack scenarios.

\subsection{Comparison with the State of the Art}
\label{sec:sota-comparison}

To benchmark its performance, we compare against a broad range of contemporary approaches on RA, RM, OULU-NPU, RY, and SiW-Mv2 datasets using standard evaluation metrics.

\subsubsection{Replay-Attack (RA)}

\begin{table}[t] \centering \caption{Replay-Attack (RA): comparison of test HTER (\%) sorted from highest (worst) to lowest (best).} \label{tab:comp_RA} \resizebox{\linewidth}{!}{ \begin{tabular}{l c c} \hline \textbf{Method} & \textbf{Year} & \textbf{HTER (\%)} \\ \hline EfficientNet-B0 \cite{huszar2024securing} & 2024 & 36.88 \\ InceptionV4 \cite{koshy2020enhanced} & 2020 & 13.54 \\ 3D ConvNet \cite{giurato2023real} & 2023 & 11.70 \\ SCNN \cite{koshy2020enhanced} & 2020 & 7.53 \\ Multi-Block LBP \cite{gunay2023face} & 2023 & 6.98 \\ MIQF+SVM \cite{chang2022face} & 2022 & 5.38 \\ GoogLeNet+GMM \cite{fatemifar2021client} & 2021 & 3.76 \\ SfSNet \cite{pinto2020leveraging} & 2020 & 3.10 \\ VGG16+GMM \cite{fatemifar2021client} & 2021 & 1.46 \\ MobileNet{+}Image Diffusion \cite{alassafi2023novel} & 2023 & 0.09 \\ ResNet50V2 \cite{alassafi2023fully} & 2023 & 0.03 \\ HybridNet II \cite{biswas2025deep} & 2025 & 0.02 \\ Lightweight 3D-DNN \cite{seegehalli2024lightweight} & 2024 & 0.00 \\ HaTFAS \cite{zhang2024efficient} & 2024 & 0.00 \\ HybridNet I \cite{biswas2025deep} & 2025 & 0.00 \\ Deformable Convolution \cite{ibrahim2025improving} & 2025 & 0.00 \\ Spatio-Temporal \cite{khan2025spatio} & 2025 & 0.00 \\ 
AdvSpoofGuard \cite{siddique2025advspoofguard} & 2025 & 0.00 \\ 
Dual-Branch \cite{jabbar2025knowledge} & 2025 & 0.00 \\ \textbf{CASO-PAD (Proposed)} & \textbf{2025} & \textbf{0.00} \\ \hline \end{tabular}} \end{table}

As shown in Table~\ref{tab:comp_RA}, CASO-PAD achieves perfect classification on the Replay-Attack dataset with 0.0\% HTER, matching or outperforming the most recent high-performing models such as AdvSpoofGuard~\cite{siddique2025advspoofguard}, Dual-Branch~\cite{jabbar2025knowledge}, and Deformable Convolution~\cite{ibrahim2025improving}. Earlier methods, including EfficientNet-B0~\cite{huszar2024securing} and InceptionV4~\cite{koshy2020enhanced}, show considerably higher error rates, confirming the progress made by modern lightweight architectures. The zero-error performance achieved by CASO-PAD underscores its strong feature discrimination and adaptation capabilities.

\subsubsection{Replay-Mobile (RM)}

\begin{table}[t] \centering \caption{Replay-Mobile (RM): comparison of test HTER (\%) sorted from highest (worst) to lowest (best).} \label{tab:comp_RM} \resizebox{\linewidth}{!}{ \begin{tabular}{l c c} \hline \textbf{Method} & \textbf{Year} & \textbf{HTER (\%)} \\ \hline VGG16+GMM \cite{fatemifar2021client} & 2021 & 17.21 \\ GoogLeNet+GMM \cite{fatemifar2021client} & 2021 & 13.56 \\ SMKFNS \cite{arashloo2020unseen} & 2020 & 11.88 \\ 3D ConvNet \cite{giurato2023real} & 2023 & 8.70 \\ MK-SVDD-Slim \cite{arashloo2021matrix} & 2021 & 7.60 \\ MKL \cite{arashloo2021matrix} & 2021 & 6.70 \\ InceptionV4 \cite{koshy2020enhanced} & 2020 & 5.94 \\ WA (PSO+PS) \cite{fatemifar2021particle} & 2021 & 5.85 \\ WA (GA+MMS+PS) \cite{fatemifar2022face} & 2022 & 5.12 \\ SCNN \cite{koshy2020enhanced} & 2020 & 4.96 \\ EfficientNet-B0 \cite{huszar2024securing} & 2024 & 4.62 \\ MobileNet{+}Image Diffusion \cite{alassafi2023novel} & 2023 & 1.14 \\ Lightweight 3D-DNN \cite{seegehalli2024lightweight} & 2024 & 0.45 \\ ResNet50V2 \cite{alassafi2023fully} & 2023 & 0.00 \\ Deformable Convolution \cite{ibrahim2025improving} & 2025 & 0.00 \\ Spatio-Temporal \cite{khan2025spatio} & 2025 & 0.00 \\ 
AdvSpoofGuard \cite{siddique2025advspoofguard} & 2025 & 0.00 \\ 
Dual-Branch \cite{jabbar2025knowledge} & 2025 & 0.00 \\ 
\textbf{CASO-PAD (Proposed)} & \textbf{2025} & \textbf{0.00} \\ \hline \end{tabular}} \end{table}

The results on the Replay-Mobile dataset, summarized in Table~\ref{tab:comp_RM}, follow a similar trend. CASO-PAD again attains 0.0\% HTER, positioning it among the top-performing methods such as Dual-Branch~\cite{jabbar2025knowledge} and AdvSpoofGuard~\cite{siddique2025advspoofguard}. Older handcrafted and hybrid approaches (e.g., WA~\cite{fatemifar2021particle,fatemifar2022face} and SCNN~\cite{koshy2020enhanced}) show much higher error rates between 5–12\%. These findings highlight CASO-PAD’s ability to generalize across mobile capture environments characterized by unstable illumination and varying conditions.

\subsubsection{OULU}
\begin{table}[t]
\centering
\caption{OULU-NPU (complete protocol): comparison of test performance sorted by ACER from highest (worst) to lowest (best).}
\label{tab:comp_OULU}
\resizebox{\linewidth}{!}{
\begin{tabular}{l c c}
\hline
\textbf{Method} & \textbf{Year} & \textbf{APCER / BPCER / ACER (\%)} \\ \hline
Texture (VAR) \cite{daniel2021texture} & 2021 & 14.5 / 15.0 / 14.8 \\
ED-LBP (VAR) \cite{shu2021face} & 2021 & 11.3 / 8.4 / 9.9 \\
Fake-Net (VAR) \cite{alshaikhli2021face} & 2021 & 5.4 / 6.9 / 6.2 \\
OFT (VAR) \cite{li2022face} & 2022 & 5.7 / 2.7 / 4.2 \\
UCDCN \cite{zhang2024ucdcn} & 2024 & 2.6 / 1.01 / 1.82 \\
3DLCN \cite{ning2024face} & 2024 & 1.5 / 0.5 / 1.0 \\
Spatio-Temporal \cite{khan2025spatio} & 2025 & 0.13 / 1.11 / 0.62 \\
KD+Depth \cite{jabbar2025knowledge} & 2025 & 0.28 / 0.83 / 0.56 \\
\textbf{CASO-PAD (Proposed)} & \textbf{2025} & \textbf{0.00 / 0.83 / 0.42} \\ \hline
\end{tabular}}
\end{table}

Table~\ref{tab:comp_OULU} reports comparative performance on the OULU-NPU benchmark under the complete protocol, where methods are evaluated using the standardized APCER, BPCER, and ACER metrics. CASO-PAD achieves an ACER of 0.42\%, delivering one of the best overall results among recent state-of-the-art approaches. In particular, the proposed method obtains an APCER of 0.00\%, indicating perfect rejection of attack presentations, while maintaining a competitive BPCER of 0.83\%.  

Compared with earlier handcrafted feature baselines such as Texture (VAR)~\cite{daniel2021texture} and ED-LBP (VAR)~\cite{shu2021face}, CASO-PAD reduces ACER by more than an order of magnitude. Furthermore, CASO-PAD remains highly competitive against recent deep architectures including 3DLCN~\cite{ning2024face} (ACER 1.0\%) and KD+Depth~\cite{jabbar2025knowledge} (ACER 0.56\%). These results demonstrate that CASO-PAD generalizes strongly across the diverse illumination, camera, and spoofing conditions present in OULU-NPU.

\subsubsection{ROSE-Youtu (RY)}

\begin{table}[t] \centering \caption{ROSE-Youtu (RY): comparison of test HTER/EER (\%) sorted by HTER from highest (worst) to lowest (best). Entries without reported HTER are listed last.} \label{tab:comp_RY} \resizebox{\linewidth}{!}{ \begin{tabular}{l c c} \hline \textbf{Method} & \textbf{Year} & \textbf{HTER/EER (\%)} \\ \hline 3D ConvNet \cite{giurato2023real} & 2023 & 21.30/- \\ ResNet50+GMM \cite{fatemifar2021client} & 2021 & 14.69/- \\ ViViT \cite{marais2023facial} & 2023 & 13.28/2.46 \\ EfficientNet-B0 \cite{huszar2024securing} & 2024 & 9.54/- \\ FASNet \cite{bousnina2021unraveling} & 2021 & 8.57/- \\ Fatemifar et al. \cite{fatemifar2022developing} & 2022 & 6.34/- \\ WA (PSO+PS) \cite{fatemifar2021particle} & 2021 & 5.61/- \\ WA (GA+MMS+PS) \cite{fatemifar2022face} & 2022 & 5.12/- \\ MobileNet{+}Image Diffusion \cite{alassafi2023novel} & 2023 & 4.92/4.95 \\ ResNet50V2 \cite{alassafi2023fully} & 2023 & 2.53/2.64 \\ 
AdvSpoofGuard \cite{siddique2025advspoofguard} & 2025 & 1.97/1.08 \\ 
Spatio-Temporal ($\tau{=}1$) \cite{khan2025spatio} & 2025 & 1.47/0.85 \\
CA-FAS \cite{long2024confidence} & 2024 & 1.37/- \\ Deformable Convolution \cite{ibrahim2025improving} & 2025 & 1.26/0.80 \\ 
Dual-Branch \cite{jabbar2025knowledge} & 2025 & 1.02/1.15 \\
\textbf{CASO-PAD (Proposed)} & \textbf{2025} & \textbf{0.82/0.82} \\ \hline \end{tabular}} \end{table}

Table~\ref{tab:comp_RY} reports comparative performance on the more challenging ROSE-Youtu benchmark. Here, CASO-PAD achieves an HTER of 0.82\% and an EER of 0.82\%, outperforming several recent high-performing methods including Dual-Branch~\cite{jabbar2025knowledge} (1.02/1.15) and Deformable Convolution~\cite{ibrahim2025improving} (1.26/0.80).  
Compared with earlier architectures such as ViViT~\cite{marais2023facial} or EfficientNet-B0~\cite{huszar2024securing}, the improvement margin exceeds 90\%, demonstrating substantial gains in discriminability and generalization. The results on ROSE-Youtu validate CASO-PAD’s robustness against diverse spoofing modalities and sensor variations.

\subsubsection{Spoofing in the Wild (SiW-Mv2)}

\begin{table}[t]
\centering
\caption{SiW-Mv2 (Protocol-1): comparison of test HTER/EER (\%) sorted by HTER from highest (worst) to lowest (best).}
\label{tab:comp_SiWMv2}
\resizebox{\linewidth}{!}{
\begin{tabular}{l c c}
\hline
\textbf{Method} & \textbf{Year} & \textbf{HTER/EER (\%)} \\
\hline
VGG16 \cite{simonyan2014very} & 2014 & 10.54/10.45 \\
Spatio-Temporal ($\tau{=}1$) \cite{khan2025spatio} & 2025 & 6.11/6.63 \\
MobileNetV3 large \cite{howard2019searching} & 2019 & 5.93/5.87 \\
Inceptionv3 \cite{szegedy2016rethinking} & 2016 & 5.77/5.78 \\
Depth-aug. Teacher \cite{jabbar2025knowledge} & 2025 & 5.82/6.08 \\
ResNet50V2 \cite{alassafi2023fully} & 2023 & 5.18/6.08 \\
MobileNetV3 small \cite{howard2019searching} & 2019 & 5.56/5.51 \\
DeformMobileNet \cite{ibrahim2025improving} & 2025 & 4.88/4.83 \\
EfficientNet-B0 \cite{huszar2024securing} & 2024 & 4.82/5.80 \\
\textbf{CASO-PAD (Proposed)} & \textbf{2026} & \textbf{3.11/3.13} \\
\hline
\end{tabular}}
\end{table}

Table~\ref{tab:comp_SiWMv2} reports results on the SiW-Mv2 Protocol-1 benchmark, a large-scale and challenging dataset characterized by substantial subject diversity, varied attack instruments, and realistic capture conditions. Under this protocol, CASO-PAD achieves an HTER of 3.11\% and an EER of 3.13\%, yielding the lowest error rates among the compared methods.

Compared with recent lightweight architectures such as EfficientNet-B0~\cite{huszar2024securing} (4.82/5.80) and DeformMobileNet~\cite{ibrahim2025improving} (4.88/4.83), CASO-PAD reduces HTER by approximately 35--40\%, indicating improved generalization under diverse spoof scenarios. Relative to ResNet50V2~\cite{alassafi2023fully} (5.18/6.08) and the depth-augmented teacher model~\cite{jabbar2025knowledge} (5.82/6.08), the reduction in HTER exceeds 40\%, despite relying solely on RGB input without auxiliary depth or temporal modeling.

Furthermore, CASO-PAD substantially outperforms earlier baselines such as MobileNetV3~\cite{howard2019searching} and VGG16~\cite{simonyan2014very}, where error rates remain above 5--10\%. These results demonstrate that the proposed content-adaptive spatial operator enhances discriminative capability in complex, real-world spoofing conditions while preserving a lightweight and deployment-friendly architecture.

\subsubsection{Overall Discussion}

Across all three datasets, CASO-PAD consistently matches or surpasses the strongest recent competitors while maintaining a compact and computationally efficient architecture. Its use of context-adaptive involution kernels allows the model to dynamically capture spatial dependencies, yielding robust anti-spoofing performance. These results collectively establish CASO-PAD as a reliable and scalable framework for real-world FacePAD systems.

\section{Kernel Audit}\label{sec:kernel_audit}

To better understand the adaptive behavior of the proposed model, we performed a comprehensive audit of the learned kernel functions. This analysis is conducted on RY dataset and it focuses on four complementary indicators that collectively describe the kernel’s spatial and spectral properties:

\begin{itemize}
    \item \textbf{HF/LF ratio} — the proportion of high- to low-frequency energy, where higher values indicate sharper, edge-rich responses;
    \item \textbf{Anisotropy} — the degree of orientation selectivity ($0$ indicates isotropic kernels, while larger values reflect stronger directional sensitivity);
    \item \textbf{DC offset} — the mean value of kernel weights, where values near zero correspond to center–surround or high-pass behavior without brightness bias; and
    \item \textbf{Position variance} — a measure of spatial non-stationarity, with lower values denoting greater position invariance.
\end{itemize}

\paragraph{Test-set summary.}
The evaluation on the held-out test set reveals that the involution-based head produces kernels that are nearly \textbf{zero-mean and spatially invariant} (DC offset $\approx 0.0000 \pm 0.0000$, position variance $\approx 4\times10^{-6} \pm 3\times10^{-6}$). These kernels exhibit \textbf{moderate orientation selectivity} (anisotropy $0.1832 \pm 0.0696$) and a noticeable \textbf{high-frequency preference} (HF/LF $17.58 \pm 4.65$).  

\begin{figure}[ht]
  \centering
  \includegraphics[width=\linewidth]{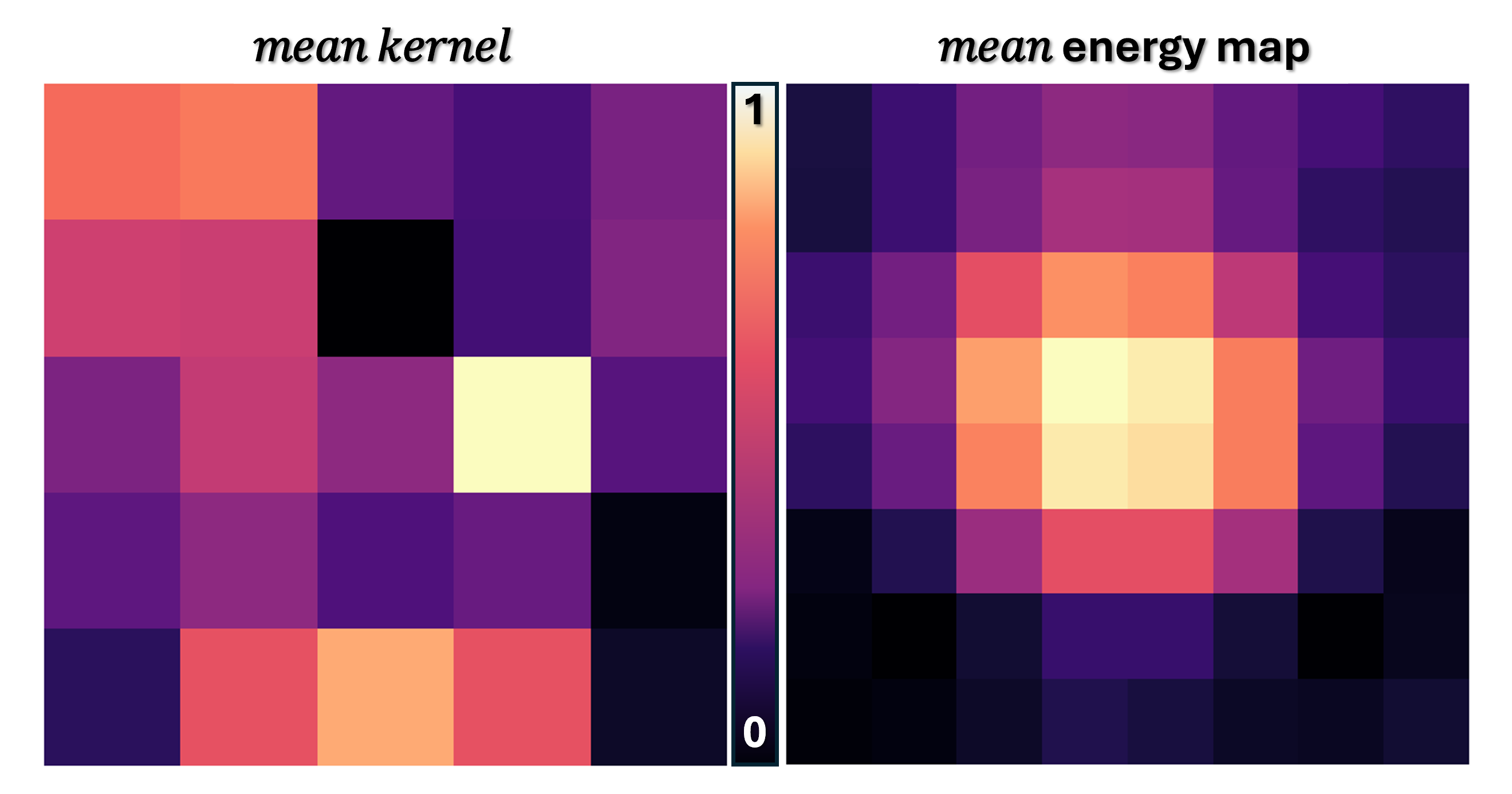}
  \caption{Kernel audit visualization (normalized to $[0,1]$). 
  \textbf{Left:} Mean kernel showing directional polarity contrast, resembling an edge-detecting filter. 
  \textbf{Right:} Mean energy distribution, radially compact and indicative of high-pass/edge-emphasizing behavior.}
  \label{fig:kernel_audit_energy}
\end{figure}

Figure~\ref{fig:kernel_audit_energy} illustrates these properties. The mean kernel (left) shows a localized polarity contrast, similar to an oriented edge detector rather than a symmetric center–surround pattern. The corresponding energy map (right) displays a radially concentrated distribution, characteristic of high-pass filtering. Together, these visualizations confirm that the model learns to enhance structural gradients and texture cues typical of genuine facial regions, while naturally suppressing smooth or redundant low-frequency content.

\begin{figure*}[t]
  \centering
  \begin{subfigure}[t]{0.47\textwidth}
    \includegraphics[width=\linewidth]{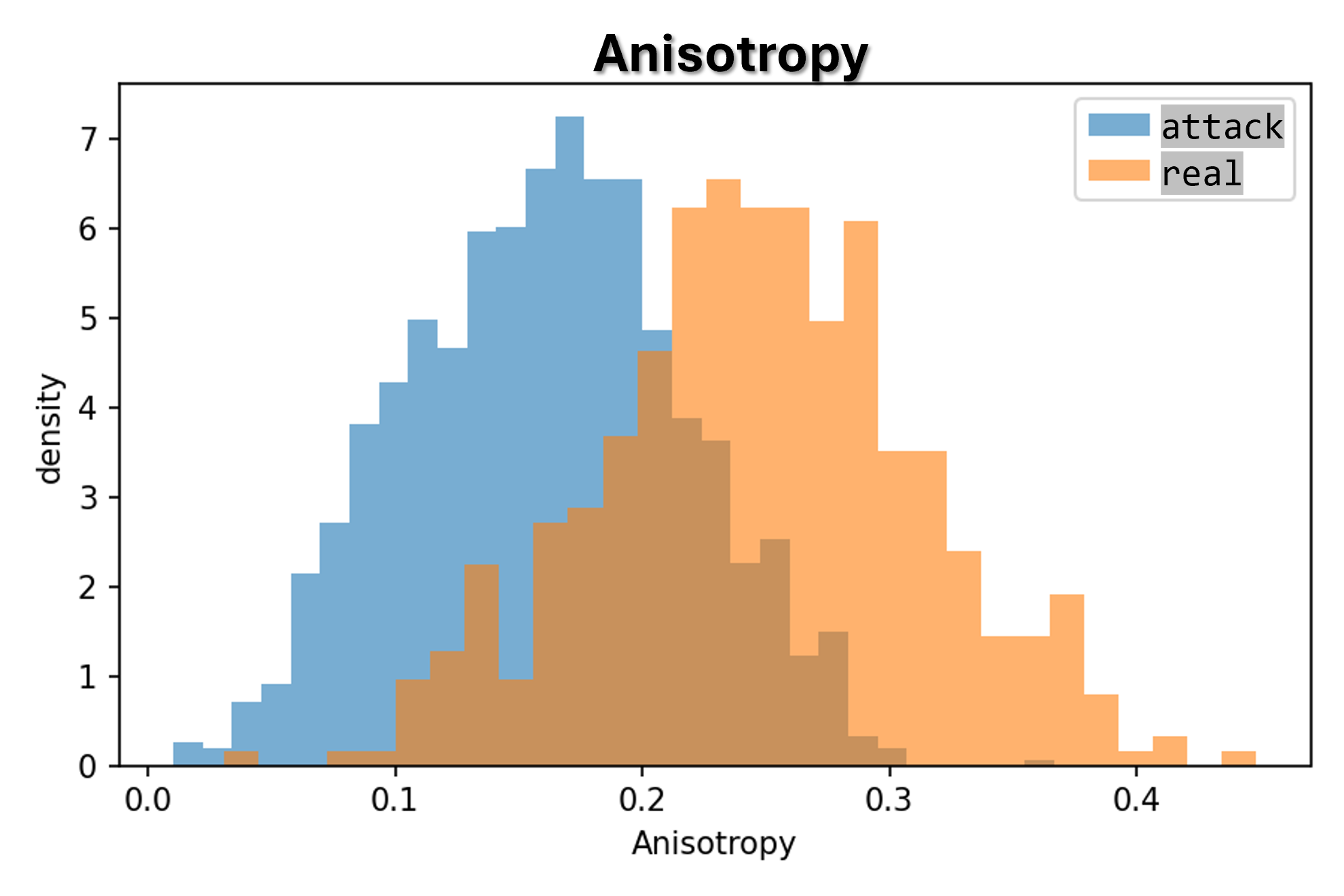}
    \caption{Anisotropy distribution}
    \label{fig:test_anisotropy_overlay}
  \end{subfigure}\hfill
  \begin{subfigure}[t]{0.47\textwidth}
    \includegraphics[width=\linewidth]{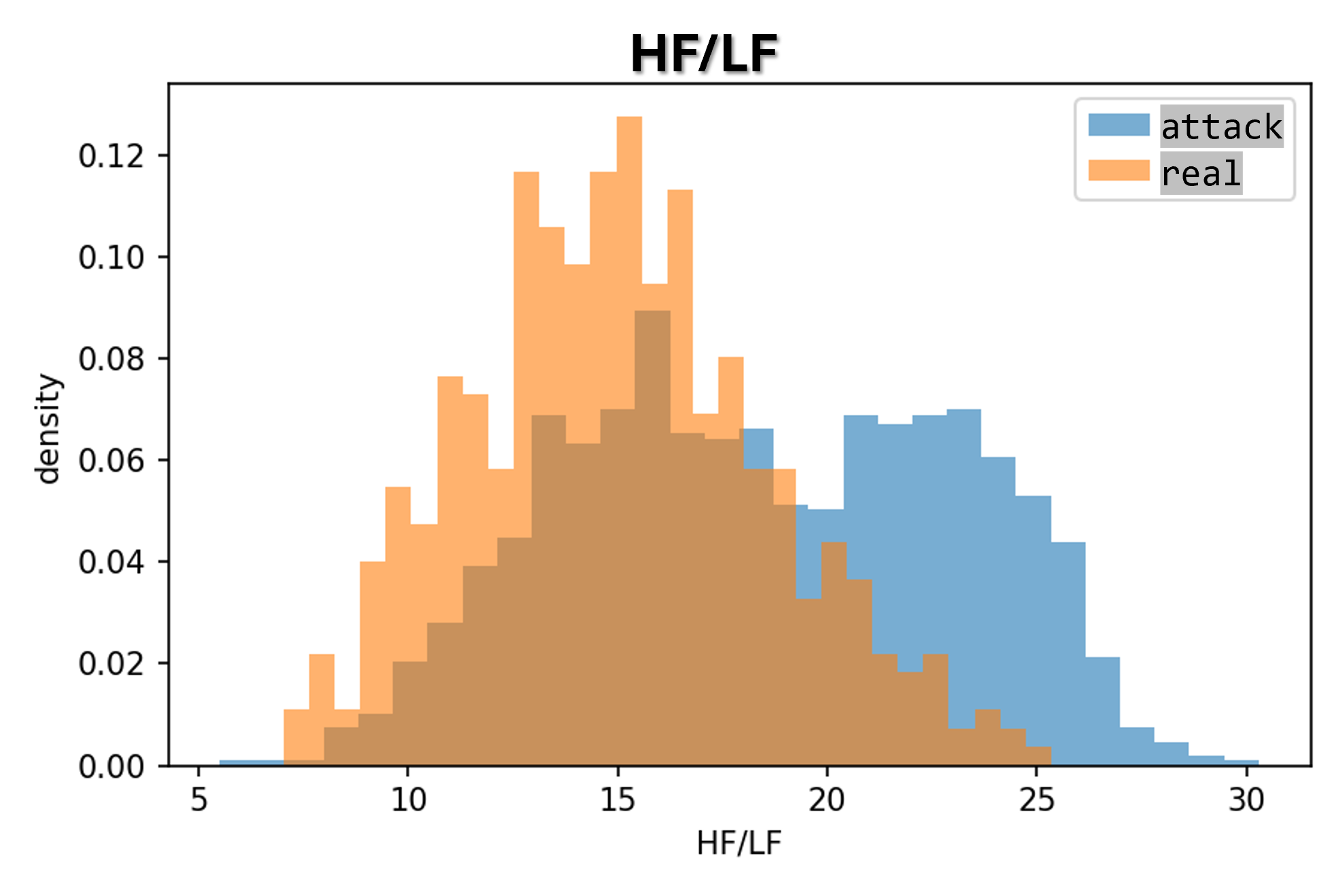}
    \caption{HF/LF distribution}
    \label{fig:test_hf_lf_overlay}
  \end{subfigure}
  \caption{Kernel audit overlays on the test set. 
  (\textbf{a}) Real faces show higher directional consistency (anisotropy), 
  while (\textbf{b}) attack samples exhibit excessive sharpness (higher HF/LF) but weaker organization.}
  \label{fig:kernel_audit_overlays}
\end{figure*}

\paragraph{Class-wise differences.}
A class-wise breakdown of the learned kernels reveals clear distinctions between genuine and spoofed inputs.  
As depicted in Figure~\ref{fig:kernel_audit_overlays}, attack samples exhibit a \textbf{higher HF/LF ratio} (18.46 vs.\ 15.02) but a \textbf{lower anisotropy} (0.161 vs.\ 0.248) than genuine faces. This indicates that spoofed frames tend to contain artificially sharp, broadband textures but lack coherent directional organization, consistent with reflections or printing artifacts.  

Cohen’s effect size further quantifies these differences: HF/LF ($d{=}+0.782$) suggests moderately higher sharpness in attacks, while anisotropy ($d{=}-1.486$) reflects a large effect favoring genuine faces with stronger directional coherence.  
The histograms in Figure~\ref{fig:kernel_audit_overlays} illustrate these tendencies: real samples cluster around higher anisotropy values, while attack samples dominate the higher HF/LF range.  
Overall, anisotropy emerges as the more discriminative feature, capturing structural regularity inherent in authentic facial geometry, whereas HF/LF primarily responds to over-sharpening and specular effects common in spoof attempts.

\paragraph{Interpretation.}
The kernel audit suggests that the involution layers act as \textit{adaptive edge-oriented spatial filters}. They enhance structured gradients aligned with genuine facial geometry while down-weighting homogeneous or specular regions often found in spoof media.  
The combination of near-zero DC bias, balanced frequency response, and moderate directional selectivity highlights the interpretability of the learned filters. In essence, the model implicitly learns a physics-consistent representation, accentuating meaningful texture and geometric cues that distinguish real from fake facial imagery.


\section{Ablation Studies}\label{sec:ablation}

This section presents a series of ablation experiments designed to evaluate the contribution of different architectural and training factors to the proposed method’s overall performance. We systematically analyze the effects of network backbone, group count in the involution head, placement of the proposed group-involution (GI) module, input image resolution, and reduction ratio. The section concludes with computational complexity analysis and qualitative interpretation using Grad-CAM visualizations.

\subsection{Effect of Network Architecture}

To understand how backbone design influences performance, two MobileNet variants were tested: MobileNetV2 and MobileNetV3-Large, each integrated with the proposed GI head. Both models were evaluated using the ROSE-Youtu dataset with identical training conditions. Key results are summarized in Table~\ref{tab:architectures}.

\begin{table}[t]
\centering
\caption{Comparison of different MobileNet backbones with proposed GI head on the ROSE-Youtu dataset (reduction ratio $=$4, groups $=$120). Results are reported as mean$\pm$std over 3 runs.}
\label{tab:architectures}
\resizebox{\linewidth}{!}{
\begin{tabular}{lcccc}
\toprule
\textbf{Model} & Params (M) & GFLOPs & Youden$_{\max}$~$\uparrow$ & HTER (\%)~$\downarrow$ \\
\midrule
MobileNetV3{+}GI & 3.635 & 0.643 & $0.984 \pm 0.004$ & $0.82 \pm 0.21$ \\
MobileNetV2{+}GI & 3.399 & 0.932 & $0.971 \pm 0.006$ & $1.43 \pm 0.34$ \\
\bottomrule
\end{tabular}}
\end{table}

MobileNetV3-Large consistently outperforms MobileNetV2, achieving higher accuracy and a lower HTER despite reduced computational cost (0.643 vs.\ 0.932 GFLOPs). The improvement can be attributed to MobileNetV3’s more expressive activation functions and squeeze-and-excitation modules, which better complement the adaptive behavior of the involution head.

\subsection{Effect of Group Count in the GI Head}

The number of groups ($G$) in the involution operator controls spatial diversity and computational load. Table~\ref{tab:groups_end_256} summarizes performance for various group counts. Performance peaks at $G{=}120$, while smaller or larger values degrade results due to underfitting or over-parameterization.

\begin{table}[h]
\centering
\caption{Ablation over group count ($G$) in the GI head (input size $256\times256$). Results are mean$\pm$std over 3 runs.}
\label{tab:groups_end_256}
\resizebox{\linewidth}{!}{
\begin{tabular}{rcccc}
\toprule
\textbf{Groups} & \textbf{Params (M)} & \textbf{GFLOPs} & \textbf{Youden $\uparrow$} & \textbf{HTER (\%) $\downarrow$} \\
\midrule
16  & 3.476 & 0.623 & 0.977$\pm$0.004 & 1.16$\pm$0.19 \\
30  & 3.497 & 0.626 & 0.977$\pm$0.010 & 1.14$\pm$0.50 \\
60  & 3.543 & 0.631 & 0.972$\pm$0.020 & 1.41$\pm$1.01 \\
\textbf{120} & \textbf{3.635} & \textbf{0.643} & \textbf{0.984}$\pm$\textbf{0.004} & \textbf{0.82}$\pm$\textbf{0.21} \\
240 & 3.818 & 0.666 & 0.965$\pm$0.020 & 1.77$\pm$0.98 \\
\bottomrule
\end{tabular}}
\end{table}

Moderate group sizes allow the model to capture sufficient spatial variation without unnecessary overhead. Excessive grouping ($G{=}240$) increases FLOPs and parameters without measurable benefit, while very low values reduce the network’s representational capacity.

\subsection{Effect of Proposed GI Placement}

To assess how the position of the GI module affects performance, two placements were evaluated: (1) at the \textit{beginning} of the MobileNetV3 backbone and (2) at the \textit{end}, right before adaptive average pooling. Results are summarized in Table~\ref{tab:placement}.

\begin{table}[h]
\centering
\caption{Effect of GI block placement in MobileNetV3 (groups$=120$). Values are mean$\pm$std over 3 runs. GFLOPs measured on $256\times256$ input.}
\label{tab:placement}
\resizebox{\linewidth}{!}{
\begin{tabular}{lcccc}
\toprule
Placement & Params (M) & GFLOPs & Youden$_{\max}$~$\uparrow$ & HTER (\%)~$\downarrow$ \\
\midrule
Beginning & 2.975 & 0.645 & $0.968 \pm 0.014$ & $1.59 \pm 0.70$ \\
End       & 3.635 & 0.643 & $0.984 \pm 0.004$ & $0.82 \pm 0.21$ \\
\bottomrule
\end{tabular}}
\end{table}

Placing the GI at the end yields superior results, indicating that context-adaptive filtering benefits most from high-level semantic representations. Early placement offers slightly reduced computational cost but sacrifices discriminative power.

\subsection{Effect of Input Image Size}

\begin{figure*}[ht]
  \centering
  \includegraphics[width=\linewidth]{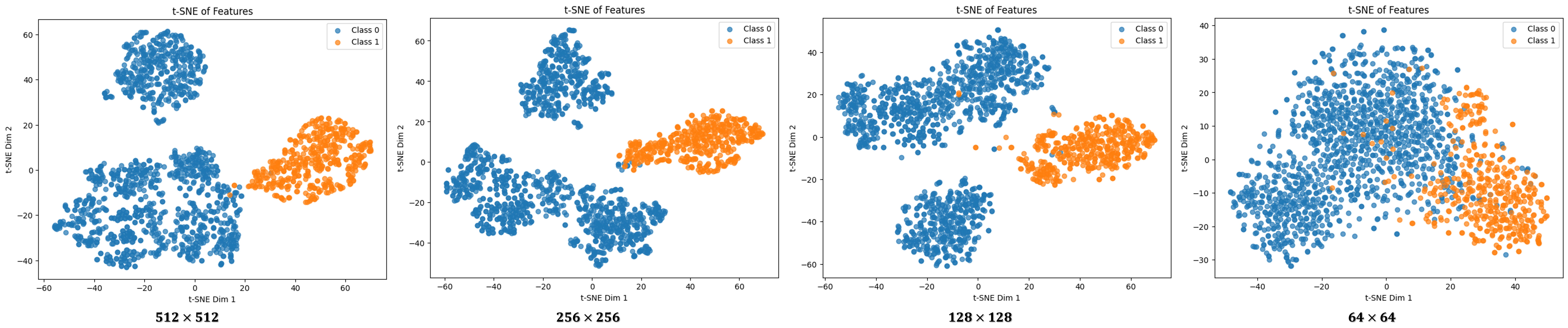}
  \caption{t-SNE embeddings of features before the classification layer across different input resolutions. Higher resolutions yield better class separation.}
  \label{fig:RY_TSNE}
\end{figure*}

To study the trade-off between input resolution and accuracy, models were trained with image sizes of $64\times64$, $128\times128$, $256\times256$, and $512\times512$. Results are reported in Table~\ref{tab:imagesize} and illustrated in Figure~\ref{fig:RY_TSNE}.
The t\text{-}SNE plots show a clear resolution-separation trend: at \(64{\times}64\) the bona fide and spoof clusters overlap notably; at \(128{\times}128\) and \(256{\times}256\) clusters become more compact with wider margins; at \(512{\times}512\) separation is strongest, matching the best EER/HTER. 

\begin{table}[h]
    \centering
    \caption{Performance for different input image sizes on the ROSE-Youtu dataset. Mean$\pm$std over 3 runs.}
    \resizebox{\linewidth}{!}{
    \begin{tabular}{l | c c c c}
        \hline
        \textbf{\emph{\diagbox[width=12em]{Metric}{Image Size}}} &
        $\mathbf{512\times512}$ & $\mathbf{256\times256}$ & $\mathbf{128\times128}$ & $\mathbf{64\times64}$ \\
        \hline
        Accuracy (\%) & 98.97$\pm$0.43 & 98.90$\pm$0.35 & 98.15$\pm$1.01 & 93.48$\pm$1.17 \\
        AUC-ROC & 0.9994$\pm$0.0003 & 0.9995$\pm$0.0003 & 0.9971$\pm$0.0033 & 0.9724$\pm$0.0063 \\
        EER (\%) & 0.90$\pm$0.36 & 0.82$\pm$0.09 & 1.61$\pm$0.80 & 8.21$\pm$1.16 \\
        HTER (\%) & 0.61$\pm$0.23 & 0.82$\pm$0.21 & 1.64$\pm$0.70 & 7.70$\pm$1.42 \\
        Youden Index (YI) & 0.988$\pm$0.005 & 0.984$\pm$0.004 & 0.967$\pm$0.014 & 0.846$\pm$0.028 \\
        Params (M) & 3.635 & 3.635 & 3.635 & 3.635 \\
        GFLOPs & 2.563 & 0.643 & 0.163 & 0.043 \\
        \hline
    \end{tabular}}
    \label{tab:imagesize}
\end{table}

Performance improves with higher resolutions, with the best results achieved at $512\times512$ (HTER 0.61\%). However, computational cost grows rapidly, from 0.043 to 2.563 GFLOPs. The $256\times256$ resolution provides a balanced trade-off, offering near-maximum accuracy with manageable complexity, making it the most practical choice for real-time applications.

\subsection{Effect of Reduction Ratio}

Table~\ref{tab:reduce_ablation_end_g120} examines the influence of the reduction ratio (\textit{reduce}) in the kernel generator of the involution head. As the bottleneck ratio increases, parameter count and FLOPs decrease, but performance follows a non-linear trend. The optimal configuration is \textit{reduce}$=4$, which delivers the highest Youden Index ($0.984\pm0.004$) and lowest HTER ($0.82\pm0.21$).

\begin{table}[h]
\centering
\caption{Effect of reduction ratio (\textit{reduce}$\in\{1,4,8\}$) in the END-placed involution~\cite{li2021involution} head ($G{=}120$). Mean$\pm$std over 3 runs.}
\label{tab:reduce_ablation_end_g120}
\resizebox{\linewidth}{!}{
\begin{tabular}{lcccc}
\toprule
Reduce & Params (M) & GFLOPs & Youden$_{\max}$~$\uparrow$ & HTER (\%)~$\downarrow$ \\
\midrule
1 & 5.775 & 0.919 & $0.965 \pm 0.028$ & $1.75 \pm 1.40$ \\
4 & 3.635 & 0.643 & $0.984 \pm 0.004$ & $0.82 \pm 0.21$ \\
8 & 3.303 & 0.600 & $0.967 \pm 0.016$ & $1.65 \pm 0.78$ \\
\bottomrule
\end{tabular}}
\end{table}

A small bottleneck (\textit{reduce}$=1$) overfits and increases computational load, while an overly large one (\textit{reduce}$=8$) restricts the kernel generator’s capacity. A moderate setting at \textit{reduce}$=4$ strikes the best balance between efficiency and expressiveness.

\subsection{Computational Complexity and Efficiency Analysis}

\begin{figure*}[ht]
    \centering
    \includegraphics[width=\textwidth]{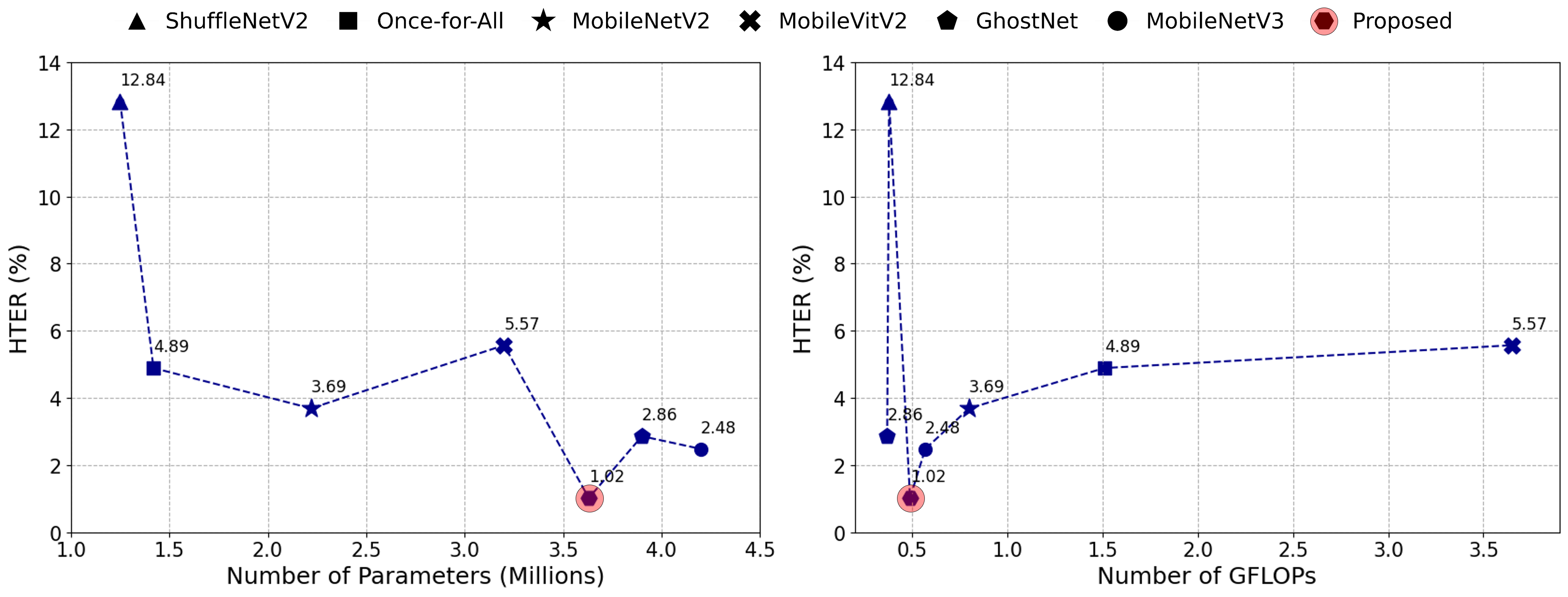}
    \caption{Trade-off comparison between model complexity and performance. (Left) HTER vs.\ number of parameters. (Right) HTER vs.\ GFLOPs.}
    \label{fig:computational_complex}
\end{figure*}

To examine computational efficiency, CASO-PAD is evaluated against representative lightweight architectures, including ShuffleNetV2~\cite{ma2018shufflenet}, Once-for-All~\cite{cai2019once}, MobileNetV2~\cite{sandler2018mobilenetv2}, MobileViTV2~\cite{mehta2022separable}, GhostNet~\cite{han2020ghostnet}, and MobileNetV3~\cite{howard2019searching}. All networks are trained and tested using identical settings with a $224\times224$ input resolution, ImageNet initialization, and consistent optimization parameters. Performance and complexity measurements are obtained under uniform benchmarking conditions to ensure fair comparison.

Figure~\ref{fig:computational_complex} illustrates the relationship between spoof detection performance and model complexity from two complementary perspectives. The parameter-based analysis (left) shows that CASO-PAD attains the lowest HTER (1.02\%) despite having only moderate model size (3.635M parameters), indicating superior parameter efficiency. Several smaller networks, such as ShuffleNetV2, exhibit noticeably higher error rates, suggesting that reduced parameter count alone does not guarantee robust liveness representation.

The GFLOPs-based view (right) further reveals that CASO-PAD achieves strong discriminative performance at low computational cost (0.48 GFLOPs). Importantly, models with comparable or even lower FLOPs do not consistently match this accuracy level, highlighting that effective architectural design and feature modeling play a more critical role than raw operation count. These observations collectively indicate that CASO-PAD improves generalization efficiency by extracting more informative representations per unit of computation rather than relying on larger networks.

\subsubsection{Edge Deployment: Jetson Orin Nano}

\begin{table}[t]
\centering
\caption{Inference latency comparison on NVIDIA Jetson Orin Nano for input resolution $224\times224$ (CUDA execution).}
\label{tab:comp_cost_orin}
\resizebox{\linewidth}{!}{
\begin{tabular}{l|c}
\hline
\textbf{Method} & \textbf{Jetson Orin Nano Latency (ms)} \\
\hline
ShuffleNetV2 \cite{ma2018shufflenet} & 22.6 $\pm$ 0.6 \\
Once-for-All \cite{cai2019once} & 49.36 $\pm$ 0.4 \\
MobileViTV2 \cite{mehta2022separable} & 28.87 $\pm$ 0.8 \\
MobileNetV2 \cite{sandler2018mobilenetv2} & 19.0 $\pm$ 0.4 \\
MobileNetV3 (large) \cite{howard2019searching} & 23.17 $\pm$ 0.7 \\
GhostNet \cite{han2020ghostnet} & 37.8 $\pm$ 0.2 \\
\textbf{CASO-PAD (Proposed)} & \textbf{25.6 $\pm$ 0.8} \\
\hline
\end{tabular}}
\\[1mm]
\begin{minipage}{\linewidth}
\scriptsize
\noindent \textit{Values denote average inference latency (mean $\pm$ std) over 100 runs on NVIDIA Jetson Orin Nano (25W mode, CUDA).}
\end{minipage}
\end{table}

Table~\ref{tab:comp_cost_orin} reports inference latency on the NVIDIA Jetson Orin Nano, reflecting realistic edge deployment conditions. While certain models exhibit slightly lower latency, these networks are associated with substantially higher HTER values (Fig.~\ref{fig:computational_complex}), indicating weaker spoof discrimination capability. CASO-PAD maintains competitive runtime performance while delivering significantly improved detection accuracy, demonstrating that modest increases in latency can be justified by notable gains in reliability and robustness.

This behavior underscores a critical practical consideration: minimal inference time alone is insufficient for secure biometric systems if accompanied by degraded liveness detection. The results confirm that CASO-PAD achieves a favorable accuracy–efficiency balance, making it suitable for resource-constrained, real-time face authentication scenarios.

\subsection{Model Interpretability via Grad-CAM}

\begin{figure*}[ht]
    \centering
    \includegraphics[width=\textwidth]{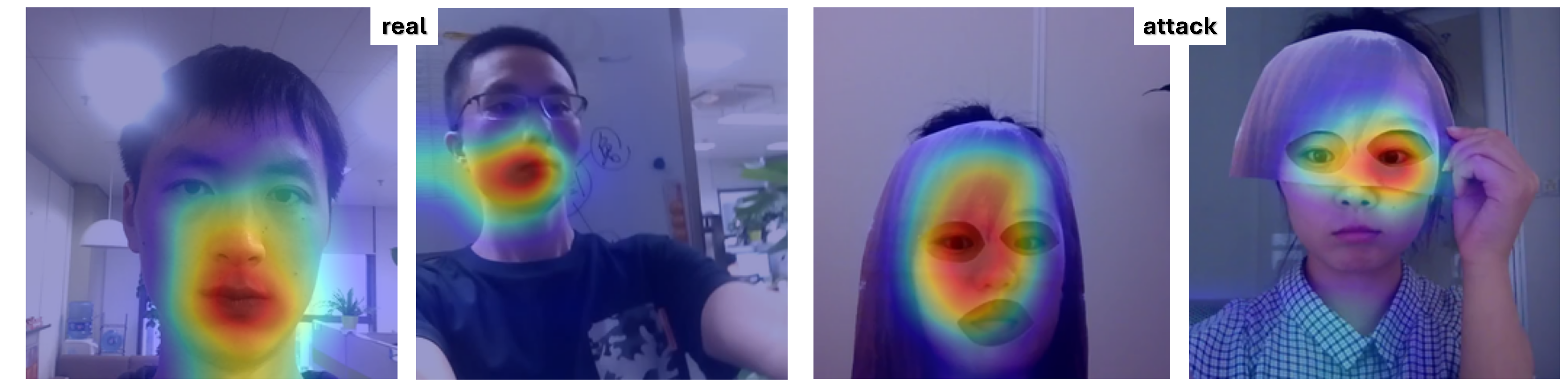}
    \caption{Grad-CAM~\cite{selvaraju2017grad} visualizations on genuine and spoof frames from the ROSE-Youtu dataset. The model focuses on discriminative cues such as eyes, lips, and mask edges (best viewed in color).}
    \label{fig:GradCAM_RY}
\end{figure*}

To interpret the model’s decision-making process, Grad-CAM heatmaps were generated for both genuine and spoof samples (Figure~\ref{fig:GradCAM_RY}). The activation maps show that the network emphasizes meaningful facial regions, such as skin texture, eyes, and lips, while also focusing on artifacts such as mask and print borders in spoofed ones. This behavior validates the discriminative nature of the learned features, illustrating that CASO-PAD effectively leverages spatial and textural cues associated with facial liveness.

\section{Conclusion}\label{sec:conclusion}

This paper introduced CASO-PAD, a lightweight Face Presentation Attack Detection (FacePAD) model that integrates content-adaptive spatial operators (involution) into a MobileNetV3 backbone. By replacing selected depthwise convolutions with a group-wise, location-adaptive operator, CASO-PAD improves spatial selectivity for spoof cues while retaining mobile-class efficiency and operating in an RGB-only, single-frame setting.

Extensive experiments on Replay-Attack, Replay-Mobile, OULU-NPU, ROSE-Youtu, and SiW-Mv2 (Protocol-1) demonstrate that CASO-PAD achieves perfect or near-ceiling performance on controlled benchmarks and maintains robust accuracy under in-the-wild conditions (e.g., 3.11\% HTER and 3.13\% EER on SiW-Mv2 Protocol-1). Ablation studies further show that placing the adaptive operator near the network head and using moderate group sharing yields the best accuracy--efficiency trade-off. In addition, the kernel audit and Grad-CAM analysis provide interpretable evidence that CASO-PAD emphasizes meaningful texture and boundary cues, while Jetson Orin Nano measurements confirm its suitability for real-time edge deployment.

The implementation of CASO-PAD is publicly available at: \url{https://github.com/Shujaat123/CASO-PAD}.


\section*{Acknowledgment}
Shujaat Khan acknowledges the support from the King Fahd University of Petroleum \& Minerals (KFUPM) under Early Career Research Grant no. EC241027.


\end{document}